\pdfoutput=1
\documentclass[letterpaper, 10 pt, conference]{ieeeconf}
\IEEEoverridecommandlockouts
\usepackage{amsmath,amssymb}
\usepackage{graphicx}
\usepackage{booktabs}
\usepackage{array}
\usepackage{cite}
\usepackage{xcolor}
\usepackage{tikz}
\usetikzlibrary{arrows.meta, positioning, calc}
\usepackage[hidelinks]{hyperref}
\definecolor{cBlue}{HTML}{1971C2}   \definecolor{cBlueFill}{HTML}{D0EBFF}
\definecolor{cTeal}{HTML}{0C8599}   \definecolor{cTealFill}{HTML}{C3FAE8}
\definecolor{cGreen}{HTML}{2F9E44}  \definecolor{cGreenFill}{HTML}{D3F9D8}
\definecolor{cOrange}{HTML}{E8590C} \definecolor{cOrangeFill}{HTML}{FFD8A8}
\definecolor{cViolet}{HTML}{7048E8} \definecolor{cVioletFill}{HTML}{E5DBFF}
\definecolor{cGray}{HTML}{495057}   \definecolor{cGrayFill}{HTML}{F1F3F5}
\newcommand{\etal}{\textit{et al.}}
\newcommand{\dnb}{{\footnotesize$\downarrow$}}
\newcommand{\upb}{{\footnotesize$\uparrow$}}


\title{\LARGE \bf Learning a Food World Model That Predicts the Next State\\ from Production Data}

\author{Inkyu~Sa,~\IEEEmembership{Senior~Member,~IEEE,} Gaurav~Kuppa, Steve~Van~der~Hoeven,\\
Konstantin~Stulov, Rajat~Bhageria, and~Luis~Rayas%
\thanks{The authors are with Chef Robotics, San Francisco, CA, USA. Corresponding author:
Luis~Rayas ({\tt\small luis@chefrobotics.ai}).}}

\newcommand{\lesson}[1]{\smallskip\noindent\textit{#1.}\ }

\begin{document}
\bstctlcite{IEEEexample:BSTcontrol}

\maketitle
{\let\thefootnote\relax\footnotetext[0]{This work has been submitted to the IEEE for possible publication. Copyright may be transferred without notice, after which this version may no longer be accessible.}}
\thispagestyle{empty}
\pagestyle{empty}

\begin{abstract}
Food assembly robots pick ingredients from a pan, and every pick must deliver a target weight. We learn a world model of food picking from 61,773 production picks, labeled automatically by a camera and a scale (self-supervised learning). From the pre-pick heightmap of the pan on a 2.5\,mm grid, the commanded pick action, the food and the utensil, the model predicts the post-pick heightmap and the picked mass with uncertainty. We evaluate the mass prediction on 12,294 held-out picks of nine foods. The production controller keeps each pick near its commanded target. The model's root-mean-square error (RMSE) of 10.93\,g is 9\% lower than that of the commanded target or of a per-food mean. This RMSE is 24\% lower than that of the controller's internal mass estimate. Gradient-boosted trees on 20 hand-crafted features have a 0.2\,g lower RMSE, but they predict only the mass, not the next state that planning and policy learning need. Using the predicted uncertainty, we decline the most uncertain picks and answer 89.4\% of picks with an RMSE of 10.28\,g. The prediction barely responds to the commanded volume and depth, because they vary little in closed-loop production. The model also fails without warning at target weights not seen in training. Future directions include action-primitive conditioning and a more general model across foods and utensils. We give recommendations for learning world models from production data.
\end{abstract}


\section{Introduction}
\label{sec:intro}

On meal assembly lines, a robot station picks an ingredient from a pan with a utensil and places it into a tray. Every pick must deliver a target weight, such as 167\,g of rice, within a tolerance that customers set in percent. A heavy portion gives food away, and a light one fails the meal specification.

\begin{figure}[!t]
\centering
\includegraphics[width=\linewidth]{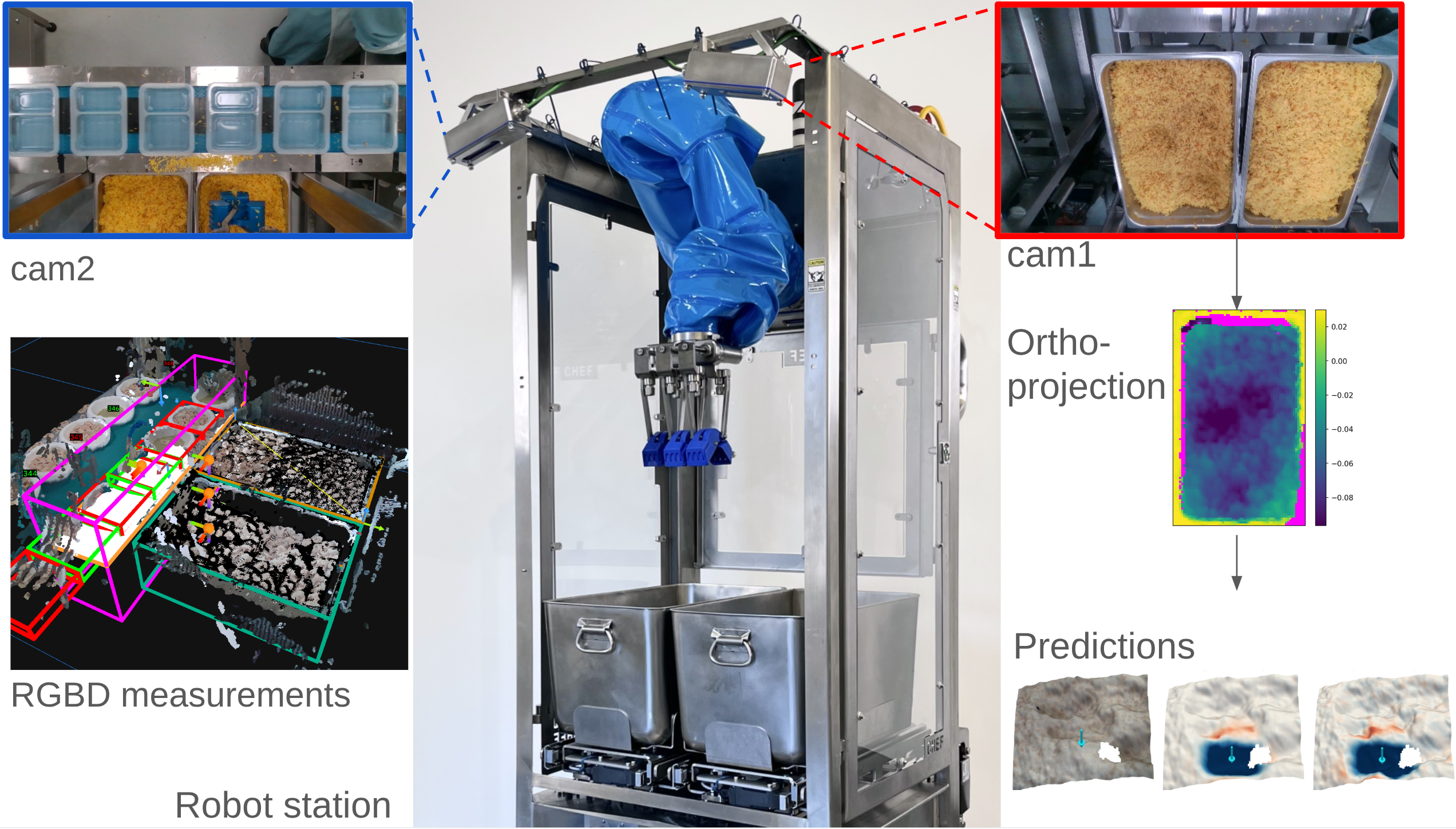}
\caption{The robot station used in this work. The model's input is the pre-pick heightmap of the pan from the top camera (cam1, ortho-projection), with meta information such as the food and the utensil ID.}
\label{fig:front}
\end{figure}

The pan changes under the robot. Each pick leaves a crater, food slides into it, and the next pick meets a different surface (Fig.~\ref{fig:front}). Foods also differ: rice holds a crater, beans in sauce flow back, and broccoli pieces roll and fall. Production controllers convert the target weight into a volume with a density from recent scale readings, then choose the depth where the heightmap says the utensil captures that volume. The density times the volume at that depth is the controller's \emph{internal estimate}. Picked masses stay near the \emph{commanded target}.

Learning a policy by trial and error in production is slow and costly. Granular and compliant food is hard to model analytically or simulate~\cite{clarke2019thesis,shi2022robocraft}. A \emph{world model}~\cite{ha2018worldmodels}, a learned model of what an action does to the environment, offers another route. Every production pick is a free training example~\cite{pinto2016supersizing,levine2018handeye}: a camera sees the surface before and after the pick, and a scale under the pan weighs the removed food. Such a model could estimate each portion before the scale settles and later help choose the pick, if it is accurate and responds to the action.

We built such a model (Fig.~\ref{fig:front}) and evaluated its prediction of the picked mass with robot inputs against the commanded target and three other baselines. Our contributions are:
\begin{itemize}
\item \emph{A food world model learned from production data:} to our knowledge, the first for food picking. From the pre-pick heightmap, it predicts the post-pick heightmap and the picked mass with uncertainty for nine foods with nonlinear, hard-to-model dynamics.
\item \emph{An evaluation on real production data against four baselines:} 9\% lower mass root-mean-square error (RMSE) than the commanded target or the mean of each food, and 24\% lower than the controller's internal estimate. Gradient-boosted trees have a 0.2\,g lower RMSE but predict only the mass, not the next state.
\item \emph{A simple, efficient architecture for deployment:} a 2.3\,M-parameter U-Net that also estimates the uncertainty of each prediction, trained with a faithful heteroscedastic loss.
\end{itemize}

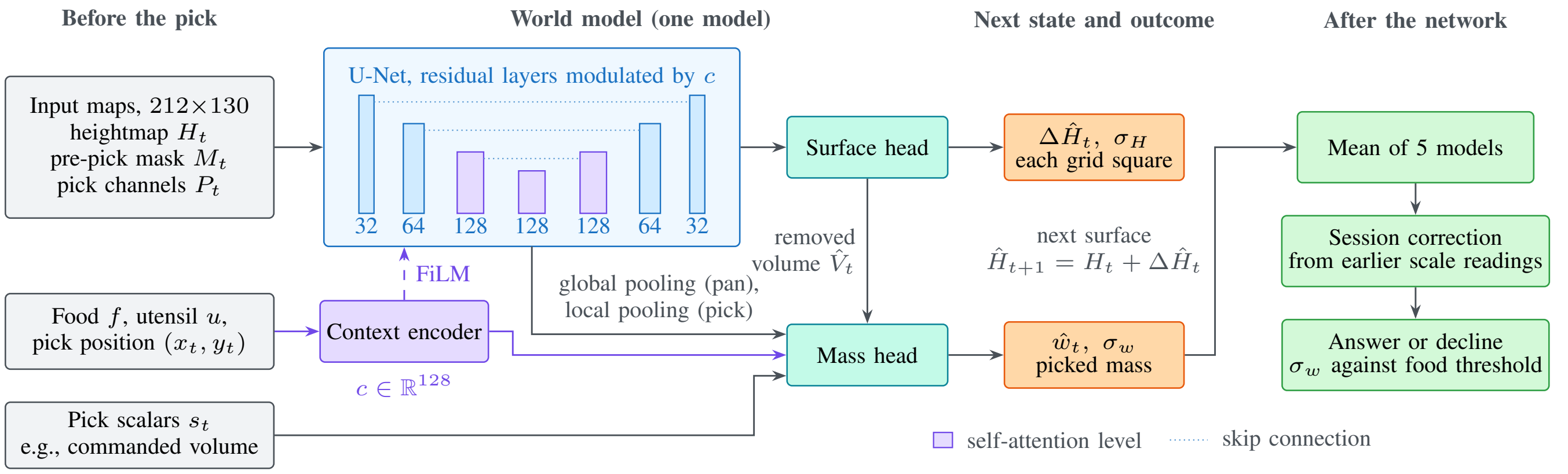
\begin{figure*}[!t]
\centering
\resizebox{\textwidth}{!}{\begin{tikzpicture}[
  font=\scriptsize,
  >={Stealth[length=1.5mm,width=1.1mm]},
  box/.style={draw=cGray, rounded corners=1.5pt, line width=0.45pt, align=center,
              inner sep=2pt},
  inbox/.style={box, fill=cGrayFill, text width=27mm},
  head/.style={box, fill=cTealFill, draw=cTeal, minimum width=17mm, minimum height=6.5mm},
  outbox/.style={box, fill=cOrangeFill, draw=cOrange, minimum width=19mm, minimum height=7mm},
  dep/.style={box, fill=cGreenFill, draw=cGreen, minimum width=25mm, minimum height=7.5mm},
  flow/.style={->, line width=0.5pt, draw=cGray},
  cflow/.style={->, line width=0.5pt, draw=cViolet},
  hdr/.style={font=\scriptsize\bfseries, text=cGray},
  note/.style={font=\scriptsize, text=cGray, align=center},
  level/.style={draw=cBlue, fill=cBlueFill, line width=0.4pt},
  alevel/.style={draw=cViolet, fill=cVioletFill, line width=0.4pt},
]
\node[hdr] at (1.45, 2.35) {Before the pick};
\node[hdr] at (6.75, 2.35) {World model (one model)};
\node[hdr] at (11.55, 2.35) {Next state and outcome};
\node[hdr] at (14.95, 2.35) {After the network};

\node[inbox, minimum height=15mm] (maps) at (1.45, 1.0)
  {Input maps, $212{\times}130$\\ heightmap $H_t$\\ pre-pick mask $M_t$\\ pick channels $P_t$};
\node[inbox, minimum height=8mm] (cond) at (1.45, -0.95)
  {Food $f$, utensil $u$,\\ pick position $(x_t, y_t)$};
\node[inbox, minimum height=7mm] (scal) at (1.45, -2.05)
  {Pick scalars $s_t$\\ e.g., commanded volume};

\node[box, fill=cBlueFill!35, draw=cBlue, minimum width=44mm, minimum height=21mm]
  (trunk) at (5.6, 1.0) {};
\node[note, text=cBlue] at (5.6, 1.74) {U-Net, residual layers modulated by $c$};
\foreach \x/\h/\w/\s/\c in {3.85/1.25/0.16/level/32, 4.35/0.95/0.22/level/64,
                            4.95/0.65/0.28/alevel/128, 5.60/0.45/0.28/alevel/128,
                            6.25/0.65/0.28/alevel/128, 6.85/0.95/0.22/level/64,
                            7.35/1.25/0.16/level/32} {
  \draw[\s] (\x-\w/2, 0.30) rectangle (\x+\w/2, 0.30+\h);
  \node[font=\scriptsize, text=cBlue] at (\x, 0.17) {\c};
}
\draw[densely dotted, draw=cBlue!70, line width=0.4pt] (3.93, 1.48) -- (7.27, 1.48);
\draw[densely dotted, draw=cBlue!70, line width=0.4pt] (4.46, 1.18) -- (6.74, 1.18);
\draw[densely dotted, draw=cBlue!70, line width=0.4pt] (5.09, 0.88) -- (6.11, 0.88);
\draw[flow] (maps.east) -- (trunk.west);

\node[box, fill=cVioletFill, draw=cViolet, minimum width=17mm, minimum height=6.5mm]
  (cenc) at (4.25, -0.95) {Context encoder};
\draw[cflow] (cond.east) -- (cenc.west);
\draw[cflow, dashed] (cenc.north) -- node[right, font=\scriptsize, text=cViolet] {FiLM}
  (cenc.north |- trunk.south);
\node[font=\scriptsize, text=cViolet, anchor=north] at (cenc.south) {$c \in \mathbb{R}^{128}$};

\node[head] (shead) at (9.15, 1.0) {Surface head};
\node[head] (mhead) at (9.15, -1.20) {Mass head};
\draw[flow] (trunk.east) -- (shead.west);
\draw[flow] (5.60, -0.05) -- (5.60, -0.98) -- node[above, note, pos=0.5] {global pooling (pan),\\ local pooling (pick)}
  (mhead.west |- 0,-0.98);
\draw[cflow] (cenc.east) -- ++(0.25, 0) |- (mhead.west |- 0,-1.20);
\draw[flow] (scal.east) -- (7.95, -2.05) |- (mhead.west |- 0,-1.42);
\draw[flow] (shead.south) -- node[left, note, align=right]
  {removed\\volume $\hat V_t$} (mhead.north);

\node[outbox] (sout) at (11.55, 1.0) {$\Delta\hat H_t,\ \sigma_{H}$\\[-1pt]each grid square};
\node[outbox] (mout) at (11.55, -1.20) {$\hat w_t,\ \sigma_{w}$\\[-1pt]picked mass};
\draw[flow] (shead) -- (sout);
\draw[flow] (mhead) -- (mout);
\node[note] at (11.55, -0.10) {next surface\\$\hat H_{t+1} = H_t + \Delta\hat H_t$};

\node[dep] (ens) at (14.95, 1.0) {Mean of 5 models};
\node[dep] (comp) at (14.95, -0.10) {Session correction\\[-1pt]from earlier scale readings};
\node[dep] (dec) at (14.95, -1.20) {Answer or decline\\[-1pt]$\sigma_w$ against food threshold};
\draw[flow] (mout.east) -- ++(0.32, 0) |- (ens.west);
\draw[flow] (ens) -- (comp);
\draw[flow] (comp) -- (dec);

\draw[alevel] (9.85, -2.18) rectangle (10.05, -2.02);
\node[font=\scriptsize, text=cGray, anchor=west] at (10.08, -2.10) {self-attention level};
\draw[densely dotted, draw=cBlue!70, line width=0.4pt] (12.35, -2.10) -- (12.75, -2.10);
\node[font=\scriptsize, text=cGray, anchor=west] at (12.78, -2.10) {skip connection};
\end{tikzpicture}}
\caption{Our model in the order of the data flow. A U-Net modulated by a context vector $c$ (food $f$, utensil $u$, pick position) reads the heightmap $H_t$, the pre-pick mask $M_t$ and the pick channels $P_t$ on a $212 \times 130$ grid. The surface head predicts $\Delta\hat H_t$ and $\sigma_H$ for every grid square. The mass head pools global and local features, adds $c$, the pick scalars $s_t$ and the removed volume $\hat V_t$, and predicts $\hat w_t$ and $\sigma_w$. Numbers under the U-Net are channel counts; violet levels carry self-attention.}
\label{fig:arch}
\end{figure*}

\section{Related Work}
\label{sec:related}

\emph{World models.} Learned models of how the environment responds to an action are used to plan or to train policies~\cite{ha2018worldmodels,hafner2019planet,hafner2025dreamerv3,wu2023daydreamer}, mostly in latent or image space. Our model predicts height in millimeters and mass in grams.

\emph{Granular material and food.} Graph networks learn granular and deformable dynamics~\cite{sanchezgonzalez2020gns,shi2022robocraft}, and image-space models suit piles~\cite{suh2021linear}. Schenck~\etal~\cite{schenck2017granular} predicted the heightmap of granular media after a scoop, Zhu~\etal~\cite{zhu2023fewshot} the scooped volume, and Clarke~\cite{clarke2019thesis} the scooped mass and the heightmap change. Clarke also chose scoops for a target mass. Takahashi~\etal~\cite{takahashi2021targetmass,takahashi2022entangled} predicted the mass of a food grasp with uncertainty, and Mori~\etal~\cite{mori2025ros2} the grasped weight of granular food. SCOOP'D~\cite{wang2025scoopd} learns scooping in simulation. Industrial portioning measures a volume and assumes a density~\cite{thongprasith2025portioning}, as the production controller does. These works use lab setups, simulation or a fixed density.

\emph{Uncertainty and declining.} Deep ensembles~\cite{lakshminarayanan2017ensembles,chua2018pets} and mean-and-variance networks~\cite{nix1994meanvariance,kendall2017uncertainty,stirn2023faithful} give predictive uncertainty. At a fixed reject rate~\cite{chow1970reject}, the optimal regression rule declines the inputs with the largest conditional variance~\cite{zaoui2020reject}.

\emph{Learning from production data.} Machine learning in production fails in ways that benchmarks hide~\cite{sculley2015debt}: inputs that exist only after the decision leak the outcome~\cite{kaufman2012leakage,kapoor2023leakage}, networks learn shortcuts in their data~\cite{geirhos2020shortcut}, and label errors in test sets reorder models~\cite{northcutt2021labelerrors}. Production pipelines guard against such defects by validating their data~\cite{breck2017mltestscore,breck2019datavalidation}.

\section{Methodology}
\label{sec:method}

\subsection{Data and Training Pairs}
\label{sec:data}

Each station has a UR5e arm, an Intel RealSense D415 depth camera about 0.8\,m above two pans, and scales under the pans. Our records hold 133,340 picks of nine foods (Table~\ref{tab:food}) from 43 robots at customer sites and our own facility over 19 months. A session is one production run of one robot on one food. We split sessions at random into training and test sessions.

A training pair holds the heightmaps before and after the pick and the picked mass, the drop in the scale reading. The pre-pick frame is the last depth frame with the arm clear of the pan. The post-pick frame is the first one after the scale settles. The data export had paired many frames with a neighboring pick's record. We re-pair each pick by the record shift that puts its crater nearest the commanded pick point on the food surface. We drop the pick if that crater lies more than 4\,cm away or if its weight is impossible. The checked records form 1,012 training sessions (61,773 picks) and 440 test sessions (15,598 picks). All results use checked picks.

\begin{figure*}[!t]
\centering
\resizebox{\textwidth}{!}{%
\includegraphics[trim=0 0 118 0, clip]{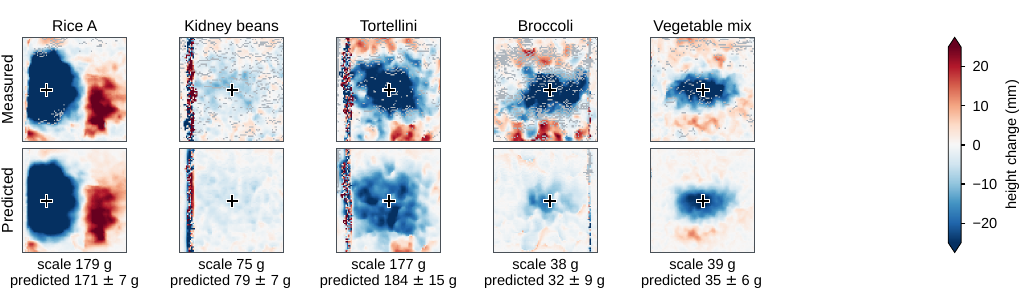}\hspace{4pt}%
\includegraphics[trim=449 0 0 0, clip]{figures/fig03_next_state_main.pdf}}
\caption{Height change of test picks, $200 \times 200$\,mm around the commanded pick ($+$), shifted to stay inside the pan. Rows: measured; predicted by one of the five models from robot inputs. Blue: food removed; red: food pushed up; gray: not measured. Columns: one test pick of each of the five foods with the most evaluation picks (selection rule in Appendix~\ref{app:surface}).}
\label{fig:state}
\end{figure*}

\subsection{Model and Training}

At pick $t$, the state is the pan surface $H_t$, a heightmap of ${212 \times 130}$ grid squares of $\delta = 2.5$\,mm over a ${530 \times 325}$\,mm pan. The mask $M_t(i) = 1$ if square $i$ has a depth reading before the pick, and 0 otherwise. The action $a_t$ is the commanded pick location and the scalars the controller sets; $k$ is the food and the utensil. Our model learns $p_\theta(\Delta H_t, w_t \mid H_t, M_t, a_t, k)$, the distribution of the surface change $\Delta H_t = H_{t+1} - H_t$ and the picked mass $w_t$, with $\Delta H_t(i) \sim \mathcal{N}(\Delta \hat H_t(i), \sigma_H^2(i))$ for each grid square $i$ and $w_t \sim \mathcal{N}(\hat w_t, \sigma_w^2)$.

The network is a U-Net~\cite{ronneberger2015unet} with a surface head and a mass head. Its residual layers are modulated (FiLM~\cite{perez2018film}) by learned food and utensil embeddings and the pick position (Fig.~\ref{fig:arch}). The surface head predicts $\Delta \hat H_t$ and $\log \sigma_H$ per square. Self-attention at the coarsest levels adds context from the whole pan. The mass head pools the bottleneck features globally, over the pan, and locally, around the pick. The mass head also takes the pick scalars and the removed volume $\hat V_t = -\delta^2 \sum_i M_t(i)\, \Delta \hat H_t(i)$ of the predicted surface change. In this way, the surface head feeds the mass head.

Each output needs an accurate mean and a spread $\sigma$ that says how far off the mean usually is. The usual Gaussian likelihood divides the gradient of the mean by $\sigma^2$, so a model can predict a large $\sigma$ on hard picks instead of fitting them~\cite{seitzer2022pitfalls}; our mass head did this and stopped learning. The faithful heteroscedastic loss~\cite{stirn2023faithful} separates the two tasks. For a residual $r = y - \hat y$ and a predicted spread $\sigma$,
\begin{equation}
\ell(r, \sigma) = \tfrac{1}{2}\, r^2 + \frac{\operatorname{sg}(r)^2}{2\sigma^2} + \log \sigma ,
\label{eq:faithful}
\end{equation}
where $\operatorname{sg}$ stops the gradient: the first term trains the mean as a plain squared error, and the last two train $\sigma$ to match the size of the residual without changing the mean. The total loss is
\begin{equation}
\mathcal{L} = \lambda_H \mathcal{L}_H + \lambda_w \mathcal{L}_w + \lambda_c \operatorname{Huber}\big((\rho_f \hat V_t - w_t)/s_w\big),
\label{eq:loss}
\end{equation}
where $\mathcal{L}_H$ is the mean of~(\ref{eq:faithful}) over the squares measured both before and after the pick, and $\mathcal{L}_w$ is~(\ref{eq:faithful}) on the standardized mass. The last term ties the picked mass to the removed volume $\hat V_t$ through a density. In this term, $\rho_f$ is one density per food (g/cm$^3$), learned with the network, and $s_w$ is the standard deviation of the training masses. The weights $\lambda_H$, $\lambda_w$ and $\lambda_c$ are hyperparameters chosen by a parameter sweep. We train five models with different seeds (AdamW~\cite{loshchilov2019adamw}, one-cycle cosine schedule, 40 epochs), each in about 2.6\,h on one GPU; Appendix~\ref{app:network} gives all settings. The model was trained with the \emph{training inputs}, whose mask marks only the squares measured both before and after the pick. A crater square often loses its reading after the pick, so the gaps of this mask mark the coming crater and leak part of the outcome.

\subsection{Model Inference}
\label{sec:robot}

\emph{Inputs.} At inference, the model receives the \emph{robot inputs}: the robot's own heightmap, the pre-pick mask $M_t$ and 3 of the 12 recorded scalars (commanded volume, tip height and fill level). The other 9 scalars, such as time stamps, are set to their training mean. A per-food offset, fitted on training predictions, removes part of the bias from this change of inputs. All results use the robot inputs. Figs.~\ref{fig:front} and~\ref{fig:state} use the stored heightmap instead.

\emph{Ensemble.} We average the five models (a model ensemble) and combine their spreads by the law of total variance~\cite{lakshminarayanan2017ensembles}.

\emph{Session correction.} Mass errors within a session are correlated, since the station, food and utensil stay the same. We therefore add to each mass prediction an exponentially weighted mean of the session's earlier residuals $w - \hat w$ (measured minus predicted mass): the \emph{session correction}.

\emph{Declining.} When $\sigma_w$ exceeds its food's threshold, the pick is \emph{declined}: the robot still picks, but only the scale checks the portion. Otherwise, the pick is \emph{answered} with an estimate.

\section{Experimental Results}
\label{sec:results}

\subsection{Setup}

\begin{table}[!t]
\caption{Data and Mass Errors per Food}
\label{tab:food}
\centering
\scriptsize
\setlength{\tabcolsep}{4pt}
\begin{tabular*}{\linewidth}{@{\extracolsep{\fill}}lrrrr@{}}
\toprule
 & & Median & \multicolumn{2}{c}{Mass RMSE (g) \dnb} \\
\cmidrule(l){4-5}
Food & Picks (sess.) & mass (g) & Ours & Mean \\
\midrule
Rice A & 4,538 (209) & 168 & \textbf{9.52} & 11.01 \\
Kidney beans & 979 (19) & 79 & \textbf{7.74} & 9.29 \\
Broccoli & 2,059 (33) & 33 & \textbf{8.89} & 9.19 \\
Vegetable mix & 1,942 (33) & 35 & \textbf{5.89} & 6.10 \\
Tortellini & 1,454 (40) & 181 & \textbf{18.04} & 19.47 \\
Tofu and carrots & 728 (27) & 36 & \textbf{4.24} & 4.73 \\
Red curry sauce & 133 (13) & 116 & \textbf{7.99} & 9.14 \\
Chicken tikka masala & 298 (26) & 239 & \textbf{25.08} & 28.24 \\
Rice B & 163 (10) & 124 & \textbf{23.09} & 23.78 \\
\midrule
All foods & 12,294 (410) &  & \textbf{10.93} & 12.06 \\
\bottomrule
\end{tabular*}\par
\vspace{2pt}
\parbox{\linewidth}{\raggedright\scriptsize Picks (sess.): evaluation picks (sessions). Median mass: median scale reading of these picks. Ours: the average of our five models; Mean: mean of each food; both with the session correction. Bold: lower error in each row. Bottom: each food in a production pan before a pick (top camera).}
\vspace{3pt}
\includegraphics[width=\linewidth]{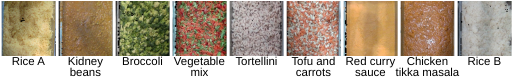}
\end{table}

\emph{Evaluation set.} We evaluate on 12,294 held-out picks of nine foods from 410 sessions (Table~\ref{tab:food}), keeping only picks whose pre-pick depth frame was taken before the pick and whose record holds the controller's internal estimate.

\emph{Mass predictors.} This section evaluates the mass; Appendix~\ref{app:surface} evaluates the surface. For each pick, the scale under the pan measures the picked mass. Before the pick, our model estimates this mass as the average of the five models (Section~\ref{sec:robot}). We compare our model with the commanded target, the \emph{mean of each food} (the average scale reading of its training picks), the internal estimate and \emph{gradient-boosted trees}~\cite{friedman2001gbm,ke2017lightgbm,grinsztajn2022tabular} on 20 pre-pick features (Appendix~\ref{app:baselines}).

\emph{Errors} are RMSEs against the scale reading, in grams. Brackets give 95\% intervals from 1,000 bootstrap resamples of whole sessions, paired between predictors.

\subsection{Picked Mass}

With the same session correction for every predictor (Table~\ref{tab:mass}), our model's RMSE of 10.93\,g is 9\% lower than that of the commanded target or the mean of each food. This RMSE is 24\% lower than that of the internal estimate. Against the mean of each food, our model is better for every food (Table~\ref{tab:food}). Gradient-boosted trees beat our model by 0.20\,g [0.08, 0.34] and are better for 8 of the 9 foods, but they predict only the mass. With declining (Section~\ref{sec:robot}), our model answers 89.4\% of picks with an RMSE of 10.28\,g, close to the 10.18\,g of the trees. The masses in Fig.~\ref{fig:state} come from one model without the session correction.

\begin{table}[!t]
\caption{Picked-Mass Error with the Session Correction}
\label{tab:mass}
\centering
\scriptsize
\setlength{\tabcolsep}{1.5pt}
\begin{tabular*}{\linewidth}{@{\extracolsep{\fill}}lrrlr@{}}
\toprule
Predictor & Ans.\ (\%) & RMSE (g) \dnb & Difference (g) [95\% CI] & $\pm$10\% \upb \\
\midrule
\multicolumn{5}{@{}l}{\emph{All picks}} \\
\quad Commanded target & 100 & 11.97 & $+1.04$ [0.89, 1.19] & 66.7 \\
\quad Mean of each food & 100 & 12.06 & $+1.13$ [0.90, 1.44] & 66.9 \\
\quad Internal estimate & 100 & 14.35 & $+3.42$ [2.46, 4.54] & 64.7 \\
\quad Gradient-boosted trees & 100 & \textbf{10.73} & $-0.20$ [$-0.34$, $-0.08$] & \textbf{71.2} \\
\quad Our model & 100 & 10.93 & ref. & 70.3 \\
\midrule
\multicolumn{5}{@{}l}{\emph{Answered picks (our model declines by $\sigma_w$)}} \\
\quad Mean of each food & 89.4 & 11.40 & $+1.12$ [0.88, 1.48] & 67.9 \\
\quad Gradient-boosted trees & 89.4 & \textbf{10.18} & $-0.10$ [$-0.23$, 0.02] & \textbf{71.9} \\
\quad Our model & 89.4 & 10.28 & ref. & 71.0 \\
\bottomrule
\end{tabular*}\par
\vspace{2pt}
\parbox{\linewidth}{\raggedright\scriptsize Evaluation set, robot inputs, session correction for every predictor. Our model: the average of the five models. Ans.: picks answered; thresholds from the other half of the test sessions (mean of 20 splits). Difference: row RMSE minus our model's on the same picks, paired over session resamples. $\pm$10\%: picks (\%) within 10\% of the scale reading. Bold: best in each part.}
\end{table}

\subsection{Uncertainty and Response to the Action}

\begin{figure}[!t]
\centering
\includegraphics[width=\linewidth]{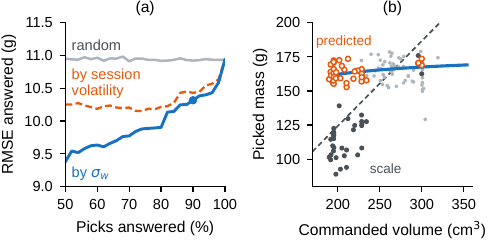}
\caption{(a)~RMSE of the answered picks, with the session correction, when declining within each food by $\sigma_w$ (blue), the session's recent volatility (orange) or at random (gray). Thresholds from the evaluated picks themselves; dot: 90\% answered. (b)~Rice~A, one utensil. Gray: 60 test picks; blue: our model's mean prediction, with volume and depth scaled by 0.7 to 1.3; dashed: mass proportional to volume. Dark dots: 37 weighed picks of a new station (33 at untrained targets); orange: predictions.}
\label{fig:unc}
\end{figure}

The spread is sized well on average: 73.3\% of the five-model errors before the correction fall within $\pm\sigma_w$ and 95.4\% within $\pm 2\sigma_w$, against 68.3\% and 95.4\% for a Gaussian. With thresholds from other test sessions, our model answers 89.4\% of picks with 10.28\,g error, against 10.92\,g for random declines (Fig.~\ref{fig:unc}(a)).

The prediction barely responds to the command. Scaling the commanded volume and depth of 60 Rice~A test picks from 0.7 to 1.3 times raises the prediction by 7.2\,g, against 99\,g for a mass proportional to volume (Fig.~\ref{fig:unc}(b)). For the 12 food and utensil pairs with at least 100 training picks, the ratio of the two rises is only 0.05 to 0.49.

On replayed Rice~A recordings of a new station, four picks at the trained 167\,g target were predicted with an RMSE of 5\,g. The other 33 picks, at untrained targets of 117 and 137\,g, were overestimated by 49\,g (29\,g with the correction). Their $\sigma_w$ was no larger than at 167\,g, and the declining rule still answered 20 of them (Appendix~\ref{app:action}).

\section{Discussion and Limitations}
\label{sec:discussion}

\emph{A world model gives more than the mass:} the trees map 20 hand-crafted features to one number. Our model predicts the post-pick heightmap (next state) and the mass, with their uncertainty, from the raw heightmap and the action, as planning and policy learning need. First, the predicted heightmap can feed later picks. Near the next pick, one model beats the mean crater (3.87 against 4.16\,mm, Appendix~\ref{app:surface}). Second, $\sigma_w$ comes with each pick, needs no history and is calibrated on average. The trees need recent readings to decline equally well (Appendix~\ref{app:uncertainty}). Third, our model scores any pick point and is differentiable in the commanded scalars, so picks can be optimized. These uses need a model that responds to the action. Ours barely does. \emph{Closed-loop data do not teach the action:} the controller kept each pick near its target, so our model learned the usual outcome of a food and utensil, not the command's effect.

\emph{Recommendations.} Check that each record's maps belong to its pick (crater position) and precede it (frame time). Build training and robot inputs with one function. Compare every output with simple baselines given the same information (Table~\ref{tab:mass}). Vary the commanded pick slightly~\cite{forssell1999closedloop} before expecting a model to learn its effect, and do not predict at untrained targets.

\emph{Limitations.} Our model was trained with an input that leaks part of the outcome, and 16\% of its training records have maps taken after the pick. The model was trained and tested only on nine foods and their utensils. Next steps are retraining on checked records with robot inputs, action-primitive conditioning and one model across foods and utensils. Our split is random by session, so training and test share robots, sites and days. Two design choices used the final test split. The model has not run on a robot.

\section{Conclusions}
\label{sec:conclusion}

On checked production picks, our model's mass prediction beats the commanded target, the mean of each food and the internal estimate. It comes within 0.2\,g of gradient-boosted trees, which predict only the mass, not the next state. However, our model fails without warning at untrained targets. Production records give labels for free, but a world model that can choose the pick needs varied actions and carefully checked records.

\clearpage
\raggedbottom
\section*{Appendices}

These appendices give the details that support the main text. RMSE is the root-mean-square error, and CI is a confidence interval. In the tables, \dnb{} marks a metric for which lower is better, and \upb{} marks a metric for which higher is better. As in the main text, ``our model'' is the average of five networks trained with different seeds. The main text calls these networks the five models. ``One network (seed 11)'' is a single one of them. The main text calls this network ``one model'' (Section~IV-B) or ``one of the five models'' (Fig.~3).

\smallskip
\noindent Appendix~\ref{app:data}: the data, the checks of the records and the evaluation set (Sections~III-A and~IV-A).\\
Appendix~\ref{app:network}: the network and all training settings (Fig.~2, Section~III-B and Eqs.~(1) and~(2)).\\
Appendix~\ref{app:inference}: the robot inputs, the ensemble, the session correction, declining and the run time (Section~III-C).\\
Appendix~\ref{app:mass}: the metrics and every mass result, also per food (Sections~IV-A and~IV-B and Tables~I and~II).\\
Appendix~\ref{app:uncertainty}: the predicted spread of the mass and declining (Section~IV-C and Fig.~4(a)).\\
Appendix~\ref{app:action}: the response to the action and the new station (Section~IV-C and Fig.~4(b)).\\
Appendix~\ref{app:baselines}: the mass baselines of Table~II in detail (Sections~IV-A and~IV-B).\\
Appendix~\ref{app:surface}: the errors of the predicted surface change and the rule that selected the picks of Fig.~3 (Section~IV-A and Fig.~3).

\useRomanappendicesfalse
\appendices

\section{Data and Checks of the Records}
\label{app:data}

\lesson{Station and split}
Each station has two Intel RealSense D415 cameras. We use only the top camera (cam1 in Fig.~1). Each depth pixel of this camera covers about 1.75\,mm of the pan. The 133,340 recorded picks come from 43 robots over 579 days (January 2025 to August 2026). A session is one uninterrupted production run of one robot on one food. A hash of the session identifier assigns each session to the training or the test split. No session is in both splits, but the splits share robots, sites and days. The test split therefore does not measure transfer to a new robot or a new site. Table~\ref{tab:S-funnel} counts the picks at each stage.

\lesson{Frames, heightmaps and labels}
The pre-pick frame is the latest depth frame before the pick with the arm clear of the pan and the scale steady. The post-pick frame is the earliest such frame after the scale settles. A frame is rejected when more than 15\% of the pan has no depth reading. The robot's code uses the same limit (Appendix~\ref{app:action}). Depth pixels are resampled onto the 2.5\,mm grid by linear interpolation. Gaps up to 15 squares (37.5\,mm) wide are filled. Wider gaps, mostly the pan walls and the arm, stay missing. The label is the drop in the scale reading: the median of a steady interval before the pick minus the median of one after the pick. On 6,139 tortellini picks of an early pilot set, the noise of this label has a median of 0.97\,g.

\lesson{Pairing and weight checks}
The data export had paired 53\% of the kept picks with the record of a neighboring pick. We re-pair each pick by shifting its maps by whole records. The chosen shift puts the crater nearest the commanded pick point. We drop the pick if that crater lies more than 4\,cm away. For 56\% of the kept picks, no crater can be located. A session rule keeps such a pick only if its shift agrees with the rest of its session. These checks remove 27\% of the usable test picks. A weight is impossible when the pick is heavier than four times the median pick of its food and utensil. This rule drops 40 of 126,910 weighed picks. For training only, a ratio screen drops 1.1\% of the picks. For each pick, the screen takes the logarithm of the ratio of mass to commanded target. The screen drops the pick when this logarithm lies more than 3.5 robust standard deviations from the median of this logarithm over the picks of the same food.

\lesson{Maps taken after the pick}
The session rule cannot detect a stretch of a session that was exported one pick late. In such a stretch, each record holds the maps of the next pick. These maps were taken after the record's own pick. A pre-pick frame is normally stamped about 3\,s before the pick time, and a late frame near the pick time. We flag a pick as late when its pre-pick frame is stamped less than 1.5\,s before the pick time. The flag holds for 20\% of the test picks (71\% of the kidney beans, 76\% of the red curry sauce) and for 16\% of the training picks. A network fed a late record sees the surface after the pick, not before the pick. Our networks learned part of this artifact: a late record shows almost no crater at its own pick. We remove late picks from the evaluation but not from training (Section~V). For the mass, late maps matter little: with the late test picks kept, our model errs by 10.77\,g, against 10.93\,g without them (Table~II).

\lesson{Evaluation set}
The evaluation set keeps the test picks whose maps were taken before the pick and whose record holds the controller's internal mass estimate (Appendix~\ref{app:baselines}). The estimate is missing where no debug record of the controller matched a pick.

\begin{table}[!t]
\caption{From Records to Evaluation Picks}
\label{tab:S-funnel}
\centering
\scriptsize
\begin{tabular*}{\linewidth}{@{\extracolsep{\fill}}lrrr@{}}
\toprule
Stage & Picks & Sessions & Robots \\
\midrule
Recorded & 133,340 & & 43 \\
Usable frames and scale reading & 107,270 & & \\
Pass the pairing and weight checks$^{a}$ & 77,371 & & \\
\midrule
Training split$^{a}$ & 61,773 & 1,012 & 40 \\
Test split & 15,598 & 440 & 36 \\
\quad Evaluation set & 12,294 & 410 & 36 \\
\quad\quad Screened (Appendix~\ref{app:mass}) & 12,159 & & \\
\quad\quad Clear frames (Appendix~\ref{app:surface}) & 11,830 & & \\
\quad\quad\quad Next pick on time (Appendix~\ref{app:surface}) & 7,778 & & \\
\bottomrule
\end{tabular*}\par
\vspace{2pt}
\parbox{\linewidth}{\raggedright\scriptsize The two splits divide the checked picks. Each indented row is a subset of the nearest less indented row above it. $^{a}$After the ratio screen; late maps kept. Evaluation set: maps taken before the pick and the controller's internal mass estimate (Appendix~\ref{app:baselines}) recorded. Screened: without the 135 picks that the ratio screen would drop. Clear: no object above the food in either frame. Next pick on time: followed directly by an on-time pick of the same session. Empty: not counted.}
\end{table}

\section{Network and Training Settings}
\label{app:network}

This appendix gives the settings behind Fig.~2 and Section~III-B.

\lesson{Input maps}
The network reads six maps of $212 \times 130$ grid squares. The first map is the standardized pre-pick height, with unmeasured squares set to zero. The second map marks the measured squares. In training, this map was the training mask, which marks the squares measured both before and after the pick (Section~III-B). At inference, this map is the pre-pick mask $M_t$ (Section~III-C). The other four maps are the pick channels $P_t$. Two of these channels are Gaussian bumps centered on the commanded pick, with standard deviations of 2\,cm and 6\,cm. The last two channels are coordinate ramps~\cite{liu2018coordconv}. These two channels hold the position of each square along the length and along the width of the pan.

\lesson{Context vector}
The food and the utensil enter as learned embeddings of 16 and 12 dimensions. The pick position enters through a small multilayer perceptron (MLP) of 8 dimensions. A three-layer MLP maps the three vectors, joined end to end, to the context vector $c \in \mathbb{R}^{128}$. For every residual layer, $c$ sets a scale and a shift of the features (feature-wise linear modulation, FiLM~\cite{perez2018film}).

\lesson{Trunk}
The trunk is the U-Net of Fig.~2. The trunk has three levels with 32, 64 and 128 channels. The grids of the three levels have $212 \times 130$, $106 \times 65$ and $53 \times 33$ squares. The bottleneck has $27 \times 17$ squares. The trunk uses group normalization~\cite{wu2018groupnorm}, and the bottleneck uses dropout of 0.15. Self-attention with four heads~\cite{vaswani2017attention} sits at the $53 \times 33$ level of the encoder, at the same level of the decoder and at the bottleneck. These three levels are the violet levels of Fig.~2.

\lesson{Surface head}
A $3 \times 3$ convolution predicts the height change $\Delta \hat H_t$ for every square. A second convolution predicts $\log \sigma_H$ from a copy of the same features with a stopped gradient. As a result, learning the spread cannot move the mean. The value of $\log \sigma_H$ is capped at 2, so $\sigma_H \le 7.4$\,mm.

\lesson{Mass head}
The mass head pools the bottleneck features twice. The global pooling covers the whole pan. The local pooling weights each square by the 6\,cm bump at the pick. The head joins the two pooled vectors with $c$, the 16 pick scalars $s_t$ and the removed volume $\hat V_t$ of the predicted surface. A two-layer MLP maps this input to the picked mass $\hat w_t$. A second MLP predicts $\log \sigma_w$ from a copy of the same input with a stopped gradient. Twelve of the pick scalars come from the pick record, and the other four encode the pick position. The recorded scalars include values that the controller set or measured, such as the commanded volume, and also time stamps and record-matching values. At inference, the robot inputs use only 3 of the 12 recorded scalars (Appendix~\ref{app:inference}).

\lesson{Size}
One network has 2,301,333 parameters, the 2.3\,M of contribution~3. About half of the parameters sit in the decoder, which serves the surface head. The mass head holds 3\% of the parameters.

\lesson{Loss}
Both heads are trained with the faithful loss of Eq.~(1)~\cite{stirn2023faithful}. In the total loss, Eq.~(2), the weights are $(\lambda_H, \lambda_w, \lambda_c) = (5, 1, 0.3)$. A parameter sweep chose these weights. The scale $s_w$ is the standard deviation of the training masses, 62.1\,g on our data. The density $\rho_f$, in g/cm$^3$, is one number per food and is learned with the network. The mass head predicts the mass directly, because predicting the product $\rho_f \hat V_t$ alone gave a 2.6\,g larger mass error on cross-validation fold 0 of the training split. Table~\ref{tab:S-settings} gives the other training settings.

\begin{table}[!t]
\caption{Training Settings}
\label{tab:S-settings}
\centering
\scriptsize
\begin{tabular*}{\linewidth}{@{\extracolsep{\fill}}ll@{}}
\toprule
Setting & Value \\
\midrule
Optimizer & AdamW~\cite{loshchilov2019adamw} \\
Weight decay & 0.05; none on normalization and bias terms \\
Batch size & 64 \\
Epochs & 40 \\
Learning-rate schedule & one-cycle cosine~\cite{smith2019superconvergence} \\
Peak learning rate & $3 \times 10^{-4}$, after a 5\% warm-up \\
Seeds & 11 to 15 (five networks) \\
GPU kernels & deterministic \\
Data augmentation & none \\
Hardware and time & one NVIDIA RTX 5090, about 2.6\,h per network \\
\bottomrule
\end{tabular*}\par
\vspace{2pt}
\parbox{\linewidth}{\raggedright\scriptsize Settings of each of the five networks. The seed sets the initialization and the order of the training data. Before the GPU kernels were made deterministic, two runs with the same seed on the same GPU differed in mass error by 0.35\,g.}
\end{table}

\lesson{Choices made on the test split}
Two design choices were made on the final test split. Both choices were scored with training inputs on all 15,598 test picks. The cosine schedule of the learning rate lowered the mass error of one network, averaged over seeds, by 0.19\,g. The ratio screen of the training picks (Appendix~\ref{app:data}) lowered the same error by 0.07\,g. The test split is therefore not independent of model selection.

\section{Inference with Robot Inputs, and Run Time}
\label{app:inference}

\lesson{Robot inputs}
The training inputs (Section~III-B) hold the 12 recorded scalars of each pick record and the heightmap stored with the record. Their mask marks the squares measured both before and after the pick. The robot inputs hold only what exists before a pick, except the fill level, which the next paragraph checks. These inputs are the pre-pick mask $M_t$, a heightmap built by the robot's own code from the recorded pre-pick depth frame, and 3 of the 12 scalars. The 3 scalars are the commanded volume, the fill level and the commanded tip height of the utensil. The robot's heightmap differs from the stored one by 0.84\,mm on average. The other 9 scalars do not exist before a pick: time stamps, a record index, record-matching values, a configuration counter and the arm height at the pre-pick frame. These 9 scalars are set to their training mean. Surface results, and the predicted height change in Figs.~1 and~3, use the stored heightmap.

The fill level is recorded after the pick, so it could leak the outcome. For 63\% of the evaluation picks, the previous pick is recorded with no refill in between. For each of these picks, the stored fill level of the previous pick is the fill level before the pick. On these picks, our model errs by 10.84\,g with the previous value and by 10.88\,g with the pick's own value: no measurable leak.

\lesson{The leak of the training mask}
A square in the new crater often loses its depth reading after the pick, so the gaps of the training mask show where the crater will be. On 523 random clear picks (Appendix~\ref{app:surface}), 17\% of the broccoli squares within 3\,cm of the commanded pick point lose their reading, against 7\% beyond 10\,cm. The crater depth is the mean height change within 3\,cm of the commanded pick point. Over all clear evaluation picks, the median measured depth is $-31.9$\,mm. The median depth predicted by one network (seed 11) is $-32.0$\,mm with the training mask, but only $-18.0$\,mm with the pre-pick mask. For most foods that hold a crater, such as broccoli, moving the gaps near the pick by 10\,cm moves the predicted crater with them. For Rice~A, whose craters keep their readings, the predicted crater stays at the commanded pick point. Table~\ref{tab:S-surface} gives the surface error with each mask.

\begin{table}[!t]
\caption{From Training Inputs to Robot Inputs, and Run Time}
\label{tab:S-robot}
\centering
\scriptsize
\setlength{\tabcolsep}{3pt}
\begin{tabular*}{\linewidth}{@{\extracolsep{\fill}}lcc@{}}
\toprule
 & One network & Our model \\
Inputs (evaluation set) & (seed 11) & (five networks) \\
\midrule
\multicolumn{3}{@{}l}{Mass RMSE (g) \dnb, no session correction} \\
Training inputs (mask from both frames) & & 10.81 \\
$+$ pre-pick mask & & 11.42 \\
$+$ robot scalars & & 11.40 \\
$+$ heightmap built by the robot's code & & 11.41 \\
$+$ per-food offset (robot inputs) & 11.42 & 11.14 \\
\midrule
Median time per pick (ms) & 38.6 & 191.7 \\
\bottomrule
\end{tabular*}\par
\vspace{2pt}
\parbox{\linewidth}{\raggedright\scriptsize Each row adds one change to the row above. The last mass row holds the full robot inputs. One network: robot inputs only. Robot scalars: 3 recorded scalars, the other 9 at their training mean. Time: 4 threads of a workstation CPU; the robot's computer was not timed.}
\end{table}

\lesson{Cost of the leak and the per-food offset}
In Table~\ref{tab:S-robot}, the pre-pick mask raises the mass error of our model from 10.81 to 11.42\,g, so the leak is worth 0.61\,g. For each food, the offset is the mean over training picks of the prediction from training inputs minus the prediction from robot inputs. The offset uses predictions only, so it reads no labels. The offset recovers 0.27 of the 0.60\,g that the robot inputs add. The offsets range from $-0.5$ to $+7.2$\,g and are largest for Rice~B and tortellini.

\lesson{Ensemble and run time}
Our model averages the five networks, $\hat w = \tfrac{1}{5}\sum_m \hat w_m$, and combines their spreads by the law of total variance~\cite{lakshminarayanan2017ensembles}, $\sigma_w^2 = \tfrac{1}{5}\sum_m \sigma_{w,m}^2 + \tfrac{1}{4}\sum_m (\hat w_m - \hat w)^2$. On four threads of a workstation CPU (Intel Core Ultra 9 285K), the median time per pick is 38.6\,ms for one network (seed 11) and 191.7\,ms for our model. One network (seed 11) errs by 0.28\,g more than our model and is five times faster.

\lesson{Session correction}
After each weighed pick, an exponentially weighted mean $\bar r_n$ of the session's residuals $w - \hat w$ is updated, with a half-life of 20 picks. The weight of each residual in this mean halves after 20 more picks. The next prediction receives the correction $b_n = \bar r_n\, n/(n+5)$, where $n$ is the number of readings seen so far. The factor $n/(n+5)$ keeps the correction small until a few readings have arrived. The half-life and the factor $n/(n+5)$ were chosen on out-of-fold predictions for training picks, never on the test split. A scale reading is ready only after the next pick starts, so pick $t$ uses the readings up to pick $t-2$.

\lesson{Declining}
A pick is answered when $\sigma_w$ is at or below the threshold of its food. Otherwise, the pick is declined: the robot still picks, but only the scale checks the portion. Thresholds are set per food because $\sigma_w$ grows with the portion size. The thresholds use only the predicted spreads, never the measured masses. For Table~II, the test sessions are split into two random halves, 20 times. Each food's threshold for one half is the 90th percentile of $\sigma_w$ in the other half, so no pick sets its own threshold. The rule is a plug-in version, per food, of the optimal reject rule for regression~\cite{zaoui2020reject}, with $\sigma_w$ in place of the true variance. Appendix~\ref{app:uncertainty} gives the results.

\lesson{Status}
Every result replays recorded picks. The surface results use the stored heightmap. The mass results with robot inputs use the heightmap from the robot's own code. The five networks are exported as one ONNX graph together with this code, but our model has not run on a robot.

\section{Picked-Mass Results in Detail}
\label{app:mass}

This appendix gives the metrics of Section~IV-A and every mass result behind Tables~I and~II. All results use the 12,294 evaluation picks and the robot inputs (Appendix~\ref{app:inference}). The screened set of Table~\ref{tab:S-funnel} drops 135 of these picks with the ratio screen of Appendix~\ref{app:data}. Screening lowers the error of every predictor by about 1.1\,g and changes the differences little.

\begin{table*}[!t]
\caption{Picked-Mass Error in Detail}
\label{tab:S-mass}
\centering
\scriptsize
\setlength{\tabcolsep}{4pt}
\begin{tabular*}{\textwidth}{@{\extracolsep{\fill}}lllrr@{}}
\toprule
Predictor & RMSE (g) [95\% CI] \dnb & Difference (g) [95\% CI] & Within 10\% (\%) \upb & MAE (g) \dnb \\
\midrule
\multicolumn{5}{@{}l}{\emph{Without the session correction, all picks}} \\
\quad Mean of each food & 12.66 [11.47, 14.17] & $+1.53$ [0.96, 2.52] & 65.9 & 8.36 \\
\quad One network (seed 11) & 11.42 [10.46, 12.52] & $+0.28$ [0.09, 0.61] & 68.7 & 7.54 \\
\quad One network (five seeds) & 11.28 $\pm$ 0.08 & & & \\
\quad Our model & 11.14 [10.25, 12.10] & ref. & 69.4 & 7.37 \\
\midrule
\multicolumn{5}{@{}l}{\emph{With the session correction, all picks}} \\
\quad Mean of each food & 12.06 [11.15, 13.13] & $+1.13$ [0.90, 1.44] & 66.9 & 8.07 \\
\quad Our model & 10.93 [10.03, 11.91] & ref. & 70.3 & 7.21 \\
\midrule
\multicolumn{5}{@{}l}{\emph{With the session correction, answered picks (89.4\%)}} \\
\quad Mean of each food & 11.40 & $+1.12$ [0.88, 1.48] & 67.9 & \\
\quad Our model & 10.28 (10.06 to 10.37) & ref. & 71.0 & \\
\bottomrule
\end{tabular*}\par
\vspace{2pt}
\parbox{\textwidth}{\raggedright\scriptsize Evaluation set, robot inputs. Our model: the average of the five networks. One network (five seeds): mean $\pm$ standard deviation of the RMSEs of the five networks. Difference: RMSE of the row minus that of our model in the same part of the table (same correction, same picks), paired over session resamples. Within 10\%: picks (\%) whose prediction lies within 10\% of the scale reading. MAE: mean absolute error. Answered picks: each food's threshold comes from the other half of the test sessions. Each value in this part is the mean over 20 splits, and parentheses give the range over the splits.}
\end{table*}

\begin{table*}[!t]
\caption{Picked-Mass Error per Food, with the Session Correction}
\label{tab:S-food}
\centering
\scriptsize
\setlength{\tabcolsep}{4pt}
\begin{tabular*}{\textwidth}{@{\extracolsep{\fill}}lrrrrlrrr@{}}
\toprule
 & & Median & \multicolumn{3}{c}{RMSE with the correction (g) \dnb} & Rel. & \multicolumn{2}{c}{Within (\%)} \\
\cmidrule(lr){4-6}\cmidrule(l){8-9}
Food & Picks (sess.) & mass (g) & Ours & Mean & Ours [95\% CI] & (\%) \dnb & $\pm\sigma_w$ & $\pm 2\sigma_w$ \\
\midrule
Rice A & 4,538 (209) & 168 & \textbf{9.52} & 11.01 & [8.63, 10.54] & 5.7 & 71.7 & 94.4 \\
Kidney beans & 979 (19) & 79 & \textbf{7.74} & 9.29 & [6.75, 8.53] & 9.8 & 72.8 & 96.5 \\
Broccoli & 2,059 (33) & 33 & \textbf{8.89} & 9.19 & [8.38, 9.53] & 27.2 & 75.0 & 97.0 \\
Vegetable mix & 1,942 (33) & 35 & \textbf{5.89} & 6.10 & [5.29, 6.65] & 16.8 & 82.0 & 96.7 \\
Tortellini & 1,454 (40) & 181 & \textbf{18.04} & 19.47 & [16.04, 20.86] & 10.0 & 65.3 & 93.1 \\
Tofu and carrots & 728 (27) & 36 & \textbf{4.24} & 4.73 & [4.04, 4.44] & 11.6 & 76.2 & 97.9 \\
Red curry sauce & 133 (13) & 116 & \textbf{7.99} & 9.14 & [5.96, 10.91] & 6.9 & 72.2 & 96.2 \\
Chicken tikka masala & 298 (26) & 239 & \textbf{25.08} & 28.24 & [20.73, 30.50] & 10.5 & 62.1 & 95.0 \\
Rice B & 163 (10) & 124 & \textbf{23.09} & 23.78 & [16.30, 30.24] & 18.6 & 75.5 & 92.0 \\
\midrule
All foods & 12,294 (410) &  & \textbf{10.93} & 12.06 & [10.03, 11.91] & 13.0 & 73.3 & 95.4 \\
\bottomrule
\end{tabular*}\par
\vspace{2pt}
\parbox{\textwidth}{\raggedright\scriptsize Evaluation set, robot inputs. Picks (sess.): evaluation picks (sessions). Median mass: median scale reading of these picks. Ours: our model, the average of the five networks. Mean: mean of each food. Both predictors have the session correction. Bold: lower RMSE in each row. Rel.: RMSE of our model divided by the median mass; for all foods, the mean of the nine values. Within $\pm\sigma_w$ and $\pm 2\sigma_w$: errors of our model (\%) before the correction that lie within one or two predicted spreads (Appendix~\ref{app:uncertainty}).}
\end{table*}

\lesson{Metrics}
The error of a pick is the predicted mass minus the scale reading. The RMSE pools the errors of all picks by root mean square, so the foods with the largest errors weigh most. Table~\ref{tab:S-food} therefore also gives the RMSE of each food.

\lesson{Mean of each food}
This mean estimator predicts one fixed mass for every pick of a food: the average scale reading of all training picks of that food, over all its utensils. The estimator reads neither the heightmap nor the command. The estimator receives the same session correction as our model, computed from its own residuals.

\lesson{Intervals}
All intervals are 95\% intervals from 1,000 bootstrap resamples of whole sessions. Differences are paired: both predictors are scored on the same resampled picks. As a check against correlation between the sessions of one robot, we also resample whole robots (36 in the evaluation set). With the session correction, the interval of the lead over the mean of each food then becomes [0.83, 1.49]\,g, instead of [0.90, 1.44]\,g.

\lesson{All picks}
Table~\ref{tab:S-mass} gives every variant. Without the session correction, our model errs by 11.14\,g, against 12.66\,g for the mean of each food. One network errs by 11.28\,g on average over the five seeds. One network (seed 11) errs by 11.42\,g. The masses under Fig.~3 come from this network, without the correction. With the correction, our model errs by 10.93\,g and the mean of each food by 12.06\,g. The percentages of the abstract and of contribution~2 are relative differences of RMSEs, rounded to a whole percent. The 9\% is $(12.06 - 10.93)/12.06$ for the mean of each food and $(11.97 - 10.93)/11.97$ for the commanded target. The 24\% is $(14.35 - 10.93)/14.35$ for the controller's internal estimate (Appendix~\ref{app:baselines}). Appendix~\ref{app:baselines} gives the errors of these baselines. The screened set of Table~\ref{tab:S-funnel} leaves out the 135 evaluation picks that the ratio screen of Appendix~\ref{app:data} would drop. Screening lowers the error of every predictor by about 1.1\,g. The differences between predictors change only slightly.

\lesson{Answered picks}
Our model declines picks by the rule of Appendix~\ref{app:inference}, with thresholds from the other half of the test sessions. Table~II scores the mean of each food and the gradient-boosted trees on the same answered picks. Each RMSE and each share within 10\% is the mean over the 20 splits. On these picks, the mean of each food errs by 11.40\,g and our model by 10.28\,g. For the interval, the 20 splits are drawn once and kept, and the thresholds are not refitted. On each session resample, the difference in RMSE is computed on the resampled answered picks of each split and averaged over the splits. The middle 95\% of these 1,000 averages gives the interval of the lead, 1.12\,g [0.88, 1.48].

\lesson{Each food}
With the correction, our model has the lower RMSE for each of the nine foods (Table~\ref{tab:S-food}). The table gives no interval for these per-food leads, only for the RMSE of our model. The leads are smallest for broccoli (8.89 against 9.19\,g) and the vegetable mix (5.89 against 6.10\,g). For these two foods, small pieces likely roll and fall. The surface then says little about how many pieces the utensil will hold. With equal weight per food, the error relative to the median pick is 13.0\% for our model, against 14.1\% for the mean of each food. Chicken tikka masala (26 sessions) and Rice~B (10 sessions) have wide intervals. A color check, followed by inspection, found 11 chicken tikka masala sessions that hold rice, out of the 105 sessions of this food in the two splits. These 11 sessions hold 44 of the 298 evaluation picks of this food. Without these sessions, the error for this food is 26.42\,g, against 25.08\,g with them. So the large error comes from the sauce itself. Four food and utensil pairs have fewer than 100 training picks, and 54 evaluation picks use them. Without these picks, and without the correction, the lead over the mean of each food falls from 1.53 to 1.08\,g.

\section{Uncertainty of the Predicted Mass and Declining}
\label{app:uncertainty}

This appendix supports Section~IV-C and Fig.~4(a). Here $\sigma_w$ is the predicted spread of our model (the average of the five networks). Errors are RMSEs in grams on the evaluation set, with robot inputs.

\lesson{Calibration} Before the session correction, 73.3\% of the errors of our model fall within $\pm\sigma_w$ and 95.4\% within $\pm 2\sigma_w$. A Gaussian predicts 68.3\% and 95.4\%. So $\sigma_w$ is sized well on average. Per food, the share within $\pm\sigma_w$ ranges from 62\% to 82\%, and the share within $\pm 2\sigma_w$ from 92\% to 98\% (Table~\ref{tab:S-food}).

\lesson{Thresholds from other sessions} With the declining rule of Appendix~\ref{app:inference}, each food's threshold comes from the other half of the test sessions. Our model then answers 89.4\% of the picks on average (88.2\% to 90.1\% over the 20 splits). With the session correction, the answered picks have an RMSE of 10.28\,g. When the same share of picks is declined at random, the RMSE is 10.92\,g.

\lesson{Thresholds from the same picks} Fig.~4(a) sets the threshold of each food from the evaluated picks themselves. Declining 10\% of the picks by $\sigma_w$ then gives 10.32\,g, and declining 20\% gives 9.90\,g. The orange curve uses a rule without $\sigma_w$. This rule declines the picks of the sessions where the recent residuals (measured minus predicted mass) of our model varied most. The rule uses the same half-life (20 picks) and reading delay as the session correction. The rule gives 10.43\,g at 90\% answered and 10.21\,g at 80\%. So when few picks are declined, the recent volatility of a session chooses the picks to decline nearly as well as $\sigma_w$. When more picks are declined, $\sigma_w$ chooses them better. We also applied this rule to the gradient-boosted trees (Appendix~\ref{app:baselines}), with the recent residuals of the trees. The trees then give 10.22\,g at 90\% answered and 9.89\,g at 80\%. These errors match or slightly beat those of our model declined by $\sigma_w$ (10.32 and 9.90\,g).

\lesson{Ranking within a food} Within each food, we compute the Spearman correlation between $\sigma_w$ and the absolute error. The median over the nine foods is only 0.10. So $\sigma_w$ ranks the errors of one food weakly. Over all foods together, the correlation is 0.37, because $\sigma_w$ has learned which foods are harder.

\section{Response to the Action and the New Station}
\label{app:action}

This appendix supports Section~IV-C, Fig.~4(b) and Section~V.

\lesson{How the response is measured}
We ran our model with robot inputs (Appendix~\ref{app:inference}) on the 12 food and utensil pairs with at least 100 training picks. For each pair, we used up to 60 on-time test picks (maps taken before the pick). Two Rice~B pairs had only 17 and 21 such picks. For each pick, we scaled the commanded volume and depth together by 0.7 to 1.3, as for a smaller or bigger scheduled pick. The reference is a mass proportional to the volume: the commanded volume times the pair's median ratio of measured mass to commanded volume. The \emph{response} is the change of the predicted mass divided by the change of the reference mass. A response of 1 means that the prediction follows the commanded volume.

\begin{figure}[!t]
\centering
\includegraphics[width=\linewidth]{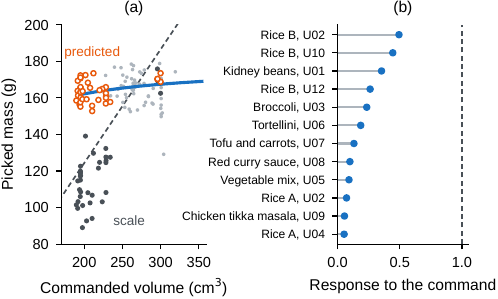}
\caption{Response of the predicted mass to the command. (a)~Rice~A with one utensil (U02), the same data as Fig.~4(b). Gray: scale readings of 60 test picks; blue: our model's mean prediction for them, with commanded volume and depth scaled by 0.7 to 1.3; dashed: mass proportional to the volume. Dark dots: scale readings of 37 weighed picks of the new station, 33 at untrained targets and 4 at the trained 167\,g target; orange circles: the predictions for these 37 picks. (b)~Response of the 12 food and utensil pairs with at least 100 training picks. The response is the change of the predicted mass from 0.7 to 1.3 times the command, divided by the change of a mass proportional to the commanded volume. A response of 1 (dashed line) means a prediction proportional to the commanded volume. Labels: the food and a utensil code.}
\label{fig:S-action}
\end{figure}

\lesson{Results}
For the Rice~A pair of the new station (U02 in Fig.~\ref{fig:S-action}), the predicted mass rose by 7.2\,g. A mass proportional to the volume would rise by 99\,g. The response of this pair is 0.07. Over the 12 pairs, the response is 0.05 to 0.49 (Fig.~\ref{fig:S-action}(b)).

\lesson{Why closed-loop data cannot teach the effect}
The production controller kept each pick near its target. The commanded depth had a standard deviation of only 2.7\,mm on one food. The training picks of the Rice~A pair of the new station were at targets of 167 and 177\,g only. These picks show almost no link between pick size and mass (a slope of $-0.06$\,g per cm$^3$ at the 167\,g target). The network therefore learned the usual outcome of a food and utensil, not the effect of the command. Small random changes of the commanded depth within the portion tolerance would add the missing variation. Such changes are the standard remedy for data recorded under closed-loop control~\cite{forssell1999closedloop}.

\lesson{Replay at a new station}
We ran our model through the robot's own code on 20 recordings from one day in late September 2026. The camera height, camera model and gripper of this station differ from those in training. The 15\% limit of the robot's code on missing depth (Appendix~\ref{app:data}) refused 32 of the 41 Rice~A picks, so we raised the limit to 20\% for the replay.

Of the 41 picks, 36 had targets of 117 and 137\,g, which this food and utensil never had in training. Of these 36 picks, 33 were weighed. Their scale readings averaged 114\,g, but the predictions stayed between 153 and 173\,g. The predictions overestimated the scale reading by 49.3\,g on average (RMSE 50.6\,g, against 11.1\,g on this pair's evaluation picks). Six of these 33 picks had frames within the 15\% limit. These six picks were overestimated by 49.0\,g, as much as the others, so the raised limit did not cause the miss. The session correction reduced the overestimate to 29.5\,g. Four weighed picks had the trained 167\,g target. Our model predicted these four picks with an RMSE of 5.0\,g. The station itself was therefore not the main problem, although four picks are few.

The spread $\sigma_w$ gave no warning: its median was 9.7\,g at the untrained targets and 10.2\,g at the trained target. The declining rule (Section~III-C) answered 20 of the 33 picks at the untrained targets. The same rule declined all four accurate picks at the trained target. A station should therefore not predict at targets outside the trained ones.

\section{Other Baselines That Use Only What Is Known before the Pick}
\label{app:baselines}

Table~II compares our model with three mass predictors besides the mean of each food. This appendix defines them and gives their errors with 95\% intervals. Section~V recommends such baselines, which receive the same information as the model. For the predicted surface change, Appendix~\ref{app:surface} gives two simple baselines: no change and the mean crater. Each predictor receives the same session correction as our model (Section~III-C).

\lesson{Predictors}
The \emph{commanded target} is the portion that the controller aimed for. The \emph{internal estimate} is the mass that the controller computes for itself: the volume that its utensil would capture at the chosen depth, times a running density from the scale readings of recent picks. The controller uses this number only internally. It is not the delivered mass. The delivered mass stays near the commanded target. The \emph{gradient-boosted trees}~\cite{friedman2001gbm} remain strong on tabular data~\cite{grinsztajn2022tabular}. We use the histogram-based regressor of scikit-learn~\cite{pedregosa2011sklearn,ke2017lightgbm} on 20 features known before the pick. Some features describe the pick: food, utensil, commanded volume, tip height and position, and the fill level before the pick. This fill level is taken from the previous recorded pick. The other features describe the pre-pick heightmap over the whole pan and within 3\,cm and 6\,cm of the commanded pick point. The number of trees (400, from candidates of up to 2,000) was chosen on 10\% of the training sessions.

\begin{table}[!t]
\caption{Other Mass Predictors, with the Session Correction}
\label{tab:S-baselines}
\centering
\scriptsize
\setlength{\tabcolsep}{2.4pt}
\begin{tabular*}{\linewidth}{@{\extracolsep{\fill}}lllr@{}}
\toprule
 & RMSE (g) \dnb & Difference (g) & Within \\
Predictor & [95\% CI] & [95\% CI] & 10\% (\%) \upb \\
\midrule
Commanded target & 11.97 [11.08, 12.96] & $+1.04$ [0.89, 1.19] & 66.7 \\
Internal estimate & 14.35 [12.91, 16.00] & $+3.42$ [2.46, 4.54] & 64.7 \\
Gradient-boosted trees & 10.73 [9.89, 11.66] & $-0.20$ [$-0.34$, $-0.08$] & 71.2 \\
Our model & 10.93 [10.03, 11.91] & ref. & 70.3 \\
\bottomrule
\end{tabular*}\par
\vspace{2pt}
\parbox{\linewidth}{\raggedright\scriptsize Evaluation set (12,294 picks), robot inputs, the same session correction for every predictor. Our model: the average of the five networks. Difference: RMSE of the row minus that of our model, paired over session resamples. Within 10\%: picks whose prediction lies within 10\% of the scale reading.}
\end{table}

\lesson{Results}
Table~\ref{tab:S-baselines} gives the errors. The 9\% for the commanded target is $(11.97 - 10.93)/11.97$, rounded to a whole percent (Appendix~\ref{app:mass} gives the 9\% for the mean of each food). The 24\% for the internal estimate is $(14.35 - 10.93)/14.35$, also rounded. The commanded target alone beats the internal estimate by 2.4\,g. The internal estimate is the weakest predictor. Its running density comes from recent picks and does not see the pan in front of the robot. The gradient-boosted trees are the strongest mass predictor. Their RMSE is 0.20\,g [0.08, 0.34] lower than that of our model ([0.07, 0.35] when whole robots are resampled). The trees are better for eight of the nine foods. Before the correction, the errors of the trees and of our model correlate at 0.94. With the correction, the average of the two predictions errs by 10.68\,g, only 0.05\,g less than the trees alone. Twenty hand-crafted features thus carry nearly all that our model extracts about the mass. For practitioners, this means: run such baselines first, with the same information and the same correction, before crediting a network with any gain.

\section{Errors of the Predicted Surface Change}
\label{app:surface}

This appendix supports Section~IV-A and Fig.~3.

\lesson{Metric}
The surface error is the RMSE of the predicted post-pick height (mm) over the squares measured in both frames, on the 5\,mm grid of the production controller. We compute it per pick and pool it by root mean square over picks. Only the interior of the pan counts: squares at least 4\,cm from the edge of the pan grid, away from the walls that the camera reads poorly. Only clear picks count, with no object (such as the utensil) above the food. We count the interior squares that lie more than 60\,mm above the median height of their frame. A pick is not clear when one frame has at least 20 such squares and the other frame has at most 2 such squares. This rule removes 3.8\% of the evaluation picks.

\lesson{Baselines and regions}
\emph{No change} predicts $\Delta \hat H = 0$. The \emph{mean crater} is the mean measured change of up to 600 checked training picks of the same food and utensil, placed at the commanded pick point. These training picks have maps taken before the pick and no object above the food (Appendix~\ref{app:data}). Besides the interior, we score the squares within 6\,cm of the commanded pick point (near pick) and within 6\,cm of the next pick's commanded point (next pick). The next pick region uses the 7,778 picks followed directly by an on-time pick. All results use robot inputs with the stored heightmap (Appendix~\ref{app:inference}).

\begin{table}[!t]
\caption{Surface Error (mm)}
\label{tab:S-surface}
\centering
\scriptsize
\setlength{\tabcolsep}{2.5pt}
\begin{tabular*}{\linewidth}{@{\extracolsep{\fill}}lrrr@{}}
\toprule
 & \multicolumn{3}{c}{Surface RMSE (mm) \dnb} \\
\cmidrule(l){2-4}
Prediction & Interior & Near pick & Next pick \\
\midrule
No change ($\Delta \hat H = 0$) & 8.84 & 25.35 & 5.56 \\
Mean crater & 4.90 & 9.87 & 4.16 \\
One network (seed 11) & 5.04 & 11.99 & 3.87 \\
Mean of the five networks & 4.81 & 11.36 & 3.74 \\
Five seeds (range) & 4.76 to 5.18 & 10.88 to 12.53 & 3.79 to 3.90 \\
\midrule
One network, training mask & 2.97 & 4.63 & n/a \\
\bottomrule
\end{tabular*}\par
\vspace{2pt}
\parbox{\linewidth}{\raggedright\scriptsize Clear evaluation picks, robot inputs with the stored heightmap. Mean of the five networks: the mean of their predicted surfaces. Five seeds: the lowest and highest of the five networks. Training mask: one network (seed 11) fed the training mask, which leaks part of the outcome. n/a: not computed.}
\end{table}

\lesson{Results}
On the interior (Table~\ref{tab:S-surface}), one network (seed 11) errs by 5.04\,mm, close to the mean crater (4.90\,mm). The paired difference is 0.14\,mm [$-0.01$, 0.29] (95\% interval) in favor of the mean crater. The interval includes zero, so the two predictions tie. The mean of the five predicted surfaces errs by 4.81\,mm. The lead of this mean over the mean crater, 0.08\,mm [$-0.06$, 0.24], is also a tie. Near the pick, the mean crater is clearly better (9.87 against 11.99\,mm). Much of this gap arises because the predicted crater is too shallow. The shallow crater comes from the leak of the training mask (Appendix~\ref{app:inference}). Fed the training mask, one network (seed 11) errs by 2.97\,mm on the interior, but the robot has only the pre-pick mask. Near the next pick, one network (seed 11) is better than the mean crater: 3.87 against 4.16\,mm, a difference of 0.30\,mm [0.18, 0.42].

\lesson{Spread}
Over the interior, 64\% of the measured changes fall within $\pm\sigma_H$ of the prediction and 86\% within $\pm 2\sigma_H$. A Gaussian gives 68\% and 95\%. Within 6\,cm of the pick, only 40\% and 62\% of the changes fall within these limits.

\lesson{Picks of Fig.~3}
The columns are the five foods with the most evaluation picks. The candidates are clear, on-time picks whose measured crater lies within 3\,cm of the commanded pick point. Each candidate also has a well-measured map and a crater at least as pronounced as the median crater of its food. Each column shows the candidate whose interior surface error is closest to the median error of its food. The predictions are from one network (seed 11) with the stored heightmap.

\section*{Acknowledgment}
The authors thank the operations and engineering teams of Chef Robotics, whose robots recorded these data.
During the preparation of this manuscript, the authors used Claude Opus 5.5 (Anthropic) for language-related assistance, including English grammar correction, sentence refinement, and general proofreading to improve readability and clarity, as well as minor LaTeX formatting adjustments. The authors have reviewed and edited the output and take full responsibility for the content of this publication.

\bibliographystyle{IEEEtran}
\bibliography{references}

\end{document}